\documentclass[runningheads]{llncs}

\usepackage[mobile]{eccv}

\usepackage{eccvabbrv}

\usepackage{graphicx}
\usepackage{booktabs}

\usepackage[accsupp]{axessibility}  %

\usepackage[pagebackref,breaklinks,colorlinks,citecolor=eccvblue]{hyperref}

\usepackage{orcidlink}
\usepackage{times}
\usepackage{epsfig}
\usepackage{amsmath}
\usepackage{amssymb}
\usepackage{float}
\usepackage{arydshln}
\usepackage{tabularx}
\usepackage{multirow}
\usepackage{makecell}
\usepackage{xspace}
\usepackage{siunitx}
\usepackage{wrapfig}
\usepackage{subcaption}

\makeatletter
\DeclareRobustCommand\onedot{\futurelet\@let@token\@onedot}
\def\@onedot{\ifx\@let@token.\else.\null\fi\xspace}

\makeatother

\usepackage[capitalize]{cleveref}
\crefname{section}{Sec.}{Secs.}
\Crefname{section}{Section}{Sections}
\Crefname{table}{Table}{Tables}
\crefname{table}{Tab.}{Tabs.}

\newcommand{\ours}{EAGLES\xspace}

\newcommand{\PreserveBackslash}[1]{\let\temp=\\#1\let\\=\temp}
\newcolumntype{C}[1]{>{\PreserveBackslash\centering}p{#1}}
\newcolumntype{R}[1]{>{\PreserveBackslash\raggedleft}p{#1}}
\newcolumntype{L}[1]{>{\PreserveBackslash\raggedright}p{#1}}

\begin{document}

\title{EAGLES: Efficient Accelerated 3D Gaussians with Lightweight EncodingS}

\author{Sharath Girish\inst{1} \and
Kamal Gupta\inst{2} \and
Abhinav Shrivastava\inst{3}
}

\authorrunning{S.~Girish et al.}

\institute{\email{sgirish@cs.umd.edu}\and
\email{kamalgupta308@gmail.com}\and
\email{abhinav@cs.umd.edu}\\
University of Maryland, College Park
}

\maketitle

\begin{figure}
    \centering
    \vspace{-1.2em}
    \includegraphics[width=0.97\textwidth]{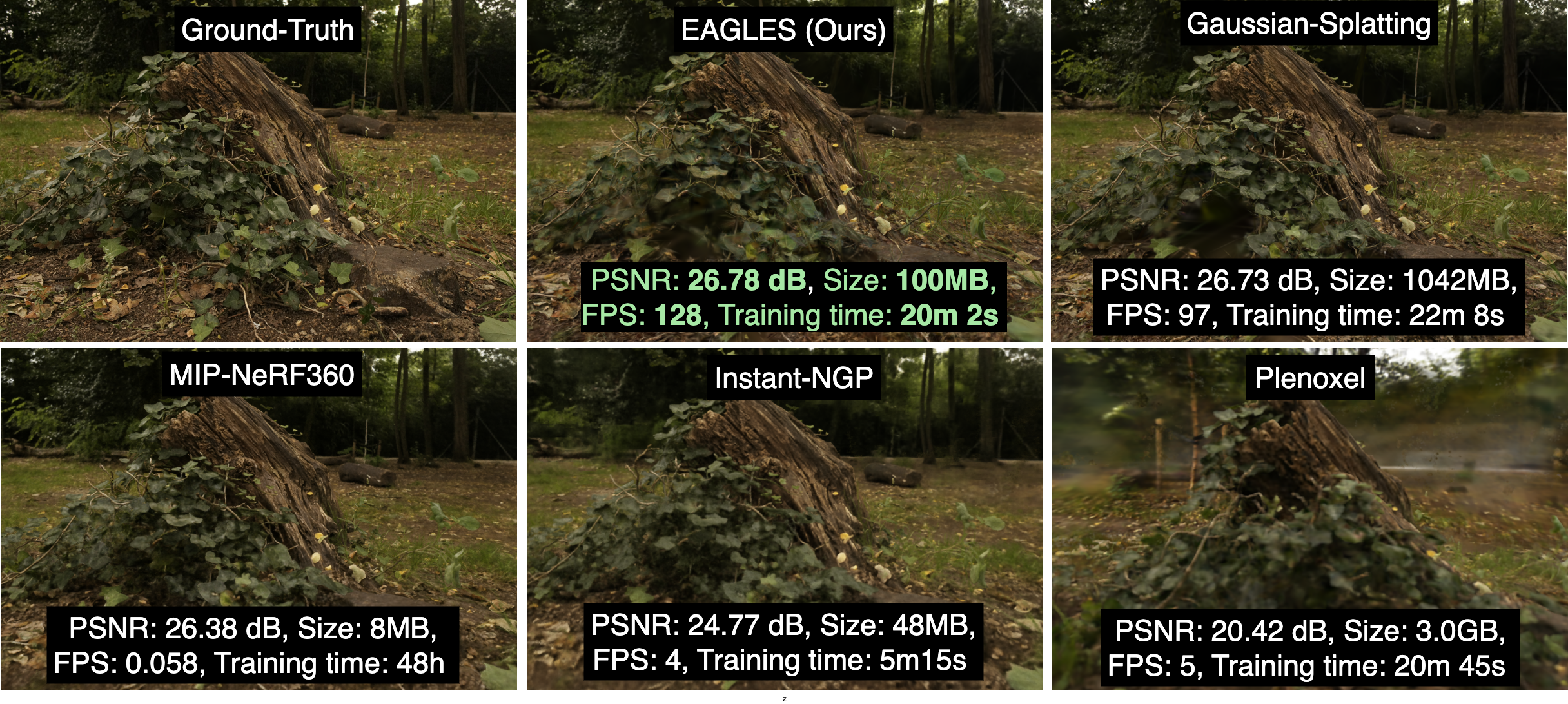}
    \vspace{-0.5em}
    \captionof{figure}{We propose \ours, a technique to significantly reduce the memory footprint of 3D-Gaussian splatting by an order of magnitude while offering further speed ups in training time and rendering FPS all while maintaining the reconstruction quality.}
    \vspace{-4em}
    \label{fig:teaser}
\end{figure}
\begin{abstract}
Recently, 3D Gaussian splatting (3D-GS) has gained popularity in novel-view scene synthesis. It addresses the challenges of lengthy training times and slow rendering speeds associated with Neural Radiance Fields (NeRFs). Through rapid, differentiable rasterization of 3D Gaussians, 3D-GS achieves real-time rendering and accelerated training.  They, however, demand substantial memory resources for both training and storage, as they require millions of Gaussians in their point cloud representation for each scene. We present a technique utilizing quantized embeddings to significantly reduce per-point memory storage requirements and a coarse-to-fine training strategy for a faster and more stable optimization of the Gaussian point clouds. Our approach develops a pruning stage which results in scene representations with fewer Gaussians, leading to faster training times and rendering speeds for real-time rendering of high resolution scenes. We reduce storage memory by more than an order of magnitude all while preserving the reconstruction quality. We validate the effectiveness of our approach on a variety of datasets and scenes preserving the visual quality while consuming 10-20$\times$ less memory and faster training/inference speed. Project page and code is available \href{https://efficientgaussian.github.io}{here}.
\end{abstract}

\section{Introduction}

Neural Radiance Fields~\cite{mildenhall2020nerf} (NeRF) have become widespread in their use as 3D scene representations achieving high visual quality by training implicit neural networks via differentiable volume rendering. They however come at the cost of high training and rendering costs. While more recent works such as Plenoxels~\cite{fridovich2022plenoxels} or Multiresolution Hashgrids~\cite{muller2022instant} have significantly reduced the training times, they are still slow to render for high resolution scenes and do not reach the same visual quality as NeRF methods such as~\cite{barron2021mip,barron2022mip}. To overcome these issues, 3D Gaussian splatting~\cite{kerbl20233d} (3D-GS) proposed an approach to learn 3D gaussian point clouds as scene representations. Unlike the slow volume rendering of NeRFs, they utilize a fast differentiable rasterizer, to project the points on the 2D plane for rendering views. They
achieve state-of-the-art (SOTA) reconstruction quality while still obtaining similar training times as the efficient NeRF variants. Through their fast tile-based rasterizer, they also achieve real-time rendering speeds at 1080p scene resolutions, significantly faster than NeRF approaches.

While 3D-GS has several advantages over NeRFs for novel view synthesis, they come at the cost of high memory usage. Each high resolution scene is represented with several millions of Gaussians in order to achieve high quality view reconstructions. Each point consists of several attributes such as position, color, rotation, opacity and scaling. This leads to representations of each scene requiring high amounts of memory for storage (${>}1$GB). The GPU runtime memory requirements during training and rendering is also much higher compared to standard NeRF methods, requiring almost 20GB of GPU RAM for several high-resolution scenes. They are thus not very practical for graphic systems with strong memory-constraints of storage or runtime memory or in low-bandwidth applications.

Our approach aims to decrease both storage and runtime memory costs while enhancing both training and rendering speeds, and maintaining view synthesis quality on par with the SOTA, 3D-GS. The color attribute, represented by spherical harmonic (SH) coefficients, and the rotation attribute, represented by covariance matrices, utilize more than 80\% of the memory cost of all attributes. Our approach significantly reduces the memory usage of each Gaussian by compressing the color and rotation attributes via a latent quantization framework. We also quantize the opacity coefficients of the Gaussians improving the optimization and leading to fewer floaters or visual artifacts in novel view reconstructions. Additionally, we propose a coarse-to-fine training strategy which improves the training stability and convergence speed while also obtaining better reconstructions. Finally, to reduce the number of redundant Gaussians resulting from frequent densification (via cloning and splitting), we utilize a pruning stage identifying Gaussians with the least influence in the full reconstruction. This further reduces the memory cost of the scene representation while improving the rendering and training speed due to faster rasterization. To summarize, our contributions are as follows:
\begin{itemize}
    \item We propose a simple yet powerful approach for compressing 3D Gaussian point clouds by quantizing per-point attributes leading to lower storage memory.
    \item We further improve the optimization of the Gaussians by quantizing the opacity coefficients and utilizing a progressive training strategy while controlling the number of Gaussians with a pruning stage.
    \item We provide ablations of the different components of our approach to show their effectiveness in producing efficient 3D Gaussian representations. We evaluate our approach on a variety of datasets achieving comparable quality as 3D-GS while being faster and more efficient.
\end{itemize}

\section{Related Work}
Neural fields or Implicit Neural Representations (INRs) have recently become a dominant representation for not just 3D objects\cite{mildenhall2020nerf,muller2022instant}, but also audio~\cite{luo2022learning,sitzmann2020implicit}, images~\cite{dupont2021coin,sitzmann2020implicit,strumpler2022implicit}, and videos~\cite{chen2021nerv,maiya2022nirvana}. Consequently, there is a big focus on improving the speed and efficiency of this line of methods. Since neural fields essentially use a neural network to represent a physical field, a number of works have been inspired by and have borrowed from the neural network compression techniques that we discuss first.

\medskip
\noindent
\textbf{Compression for neural networks}. Since the explosion of neural networks and their proliferation in the industry and applications, neural network compression and efficiency has gained a lot of attention. A typical compression scheme used for neural networks is quantization or discretization of the parameters to a smaller, finite precision and using entropy coding or other lossless compression methods to further store the parameters. While some approaches directly train binary or finite precision networks~\cite{courbariaux2015binaryconnect,li2016ternary,rastegari2016xnor,dettmers2022llm}, others attempt to quantize the network using non-uniform scalar quantization~\cite{girish2022lilnetx,zhang2018lq,banner2018post,oktay2019scalable}, or vector quantization~\cite{chen2015compressing,chen2016compressing,gong2014compressing}. Advantage of former techniques is typically cheaper setup cost and training time, however they can often result in sub-optimal network performance at the inference time. Another line of work attempt to prune the networks either during the training~\cite{lecun1990optimal,reed1993pruning,han2015learning} or in a post-hoc optimization step~\cite{frankle2018lottery,frankle2020pruning,savarese2020winning,girish2021lottery} which may require retraining the entire network. While pruning can be often a good compression strategy, these method may require a lot more training to reach a competitive performance as an unpruned network. 

\medskip
\noindent
\textbf{Compression for neural fields}. Several neural field compression approaches~\cite{sitzmann2020metasdf,tancik2021learned,strumpler2022implicit} propose a meta learning approach that learns a network on auxiliary datasets which can provide a good initialization for the downstream network. While our method can benefit from meta-learning as well, we restrict our current approach to compressing a single scene for brevity. VQAD~\cite{takikawa2022variable} propose a vector quantization for a hierarchical feature grids used in NGLOD\cite{takikawa2021neural}. Their method is able to achieve higher compression as compared to other feature-grid methods such as Instant NGP~\cite{muller2022instant} however its training can be memory intensive and it struggles to achieve the same quality of reconstructions as compared to some other NeRF variants such as MipNeRF. \cite{li2023compressing} propose a similar compression approach using voxel pruning and codebook quantization. Scalar quantization approaches~\cite{girish2023shacira,bird20213d} reparameterize the network weights with integers and apply further entropy regularization to compress the scene even further. While these approaches
require lower training memory as compared to ~\cite{takikawa2021neural}, they are sensitive to hyperparameters and the reconstruction efficacy of these approaches remain lower as compared to MipNeRF360 or Gaussian Splatting.

In this work, we show for the first time, that it is possible to compress 3D Gaussian point cloud representations which can retain high reconstruction quality with much smaller memory and higher FPS for inference.

\section{Background}
3D Gaussian splatting consists of a Gaussian point cloud representation in 3D space. Each Gaussian consists of various attributes such as the position (for mean), scaling and rotation coefficients (for covariance), opacity and color. These Gaussians represent a 3D scene and are used for rendering images from certain viewpoints by anisotropic volumetric ``splatting"~\cite{zwicker2001ewa,zwicker2001surface} of 3D Gaussians onto a 2D plane. This is done by projecting the 3D points to 2D and then using a differentiable tile-based rasterizer for blending together different Gaussians.

3D Gaussians with a mean 3D position vector $\boldsymbol{x}$ and covariance matrix $\Sigma$ can be defined as 
\begin{equation}
    G(\boldsymbol{x}) = e^{-\frac{1}{2}\boldsymbol{x}^T\Sigma^{-1}\boldsymbol{x}}
\end{equation}
The 3D covariance matrix is in turn defined using a scale matrix S (represented using a 3D scale vector $s$) and rotation matrix R (represented using a 4D rotation vector $r$) as 
\begin{equation}
    \Sigma = RSS^TR^T
\end{equation}
For a camera viewpoint with a projective transform P (world-to-camera matrix) and J as the Jacobian of the affine approximation of the projective transform, the corresponding covariance matrix projection~\cite{hoaglin1978hat} to 2D is written as:
\begin{equation}
    \Sigma' = JP\Sigma P^TJ^T
\end{equation}
The color of a pixel C is then computed using $\mathcal{N}$ Gaussian points overlapping the pixel. The points are sorted based on their depth values and blended as:
\begin{equation}
    C = \sum_{i\in\mathcal{N}}\boldsymbol{c}_i\alpha_i\prod_{j=1}^{i-1}(1-\alpha_j)
    \label{eq:raster}
\end{equation}
where $\alpha_i$ is computed by computing the 2D Gaussian at the pixel location multiplied with a scalar opacity value. The color $c_i$ of each Gaussian is then computed using spherical harmonic coefficients~\cite{seeley1966spherical}.

The Gaussians are initialized using the sparse point clouds created by Structure from Motion (SfM)~\cite{ullman1979interpretation}. Further optimization of the attributes is then done using Stochastic Gradient Descent as the rendering process is fully differentiable. For each view sampled from the training dataset, the corresponding image is projected and rasterized with the forward process explained above. The reconstruction loss is then computed by combining $\mathcal{L}_1$ with SSIM loss as 
\begin{equation}
    \mathcal{L} = (1-\lambda)\mathcal{L}_1+\lambda\mathcal{L}_\text{SSIM}
\end{equation}
with $\lambda$ set to 0.2.

Another key step in the optimization of the Gaussians is controlling the number of Gaussians. After a warmup-phase, Gaussians with a low opacity value $\alpha$ below a threshold are removed every 100 iterations. Additionally, large Gaussians (bigger than the corresponding geometry) are split while small Gaussians are cloned in order to better fit the underlying geometric shape. Only Gaussians with positional gradients above a threshold $\tau_{thresh}$ after every 100 iterations are split or cloned. 

\begin{figure*}[t]
    \centering
    \includegraphics[width=\linewidth]{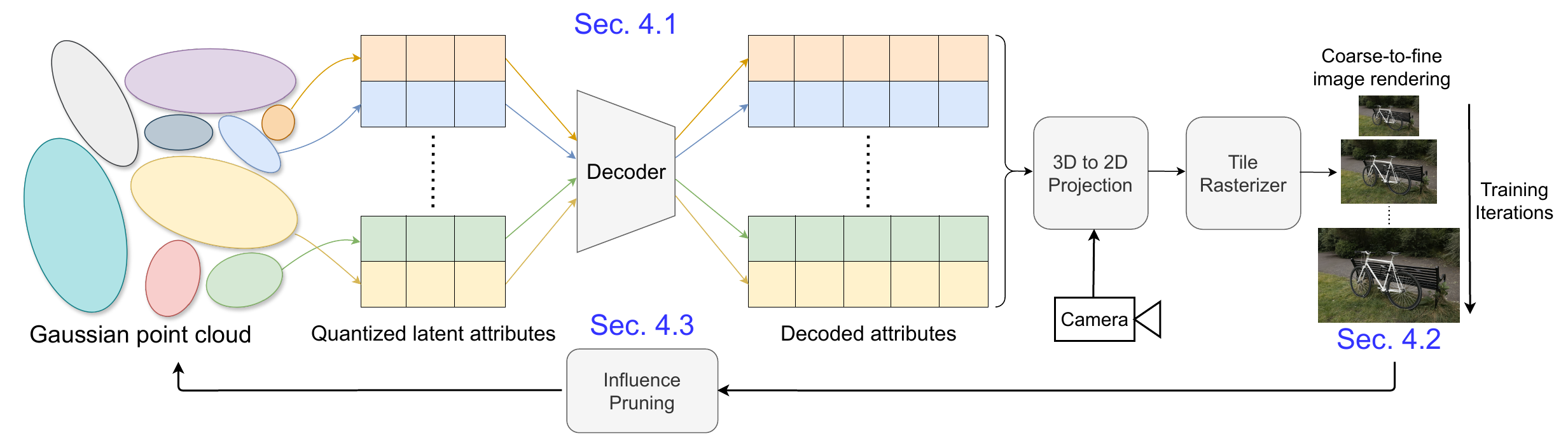}
    \vspace{-1em}
    \caption{Approach: 1) We quantize the attributes of the latents to reduce the storage memory of the Gaussians (\cref{ssec:quant}), 2) progressively train using a coarse-to-fine rendering resolution schedule to obtain higher quality reconstructions (\cref{ssec:quant} and 3) utilize a pruning stage to obtain fewer Gaussians and faster training/rendering speeds (\cref{ssec:densify}).}
    \label{fig:approach}
    \vspace{-1em}
\end{figure*}

\section{Method}
\subsection{Attribute quantization}
\label{ssec:quant}
Each Gaussian point consists of a position vector $\boldsymbol{p}\in\mathbb{R}^3$, scaling coefficient $\boldsymbol{s}\in\mathbb{R}^3$, rotation quarternion vector $\boldsymbol{r}\in\mathbb{R}^4$, opacity scalar $o\in\mathbb{R}$ and spherical harmonics coefficients $\boldsymbol{c}\in\mathbb{R}^d$, with $d=3*f^2$, where $f$ corresponds to the harmonics degree. Thus, for a degree of 4 (as is used in \cite{kerbl20233d}), the color coefficients make up more than $80\%$ of the dimensions of the full attribute vector. 3D-GS typically requires millions of Gaussians for representing the scene with high quality. However, a set of 1 million Gaussians consume around 236 MB of disk space when storing the full attribute vector with a 32-bit floating point. Thus, to reduce the memory required for storing each attribute vector, we propose to use a set of quantized representations. A visualization of the various components of our approach is provided in \cref{fig:approach}.

For any given attribute, we maintain a quantized latent vector $\boldsymbol{q}\in\mathbb{Z}^l$ with dimension $l$, consisting of integer values. We then use an MLP decoder $D: \mathbb{Z}^l{\rightarrow}\mathbb{R}^k$ to decode the latents and obtain the attributes. As quantized vectors are not differentiable, we maintain continuous approximations $\widehat{\boldsymbol{q}}$ during training and use the Straight-Through Estimator (STE) which rounds off $\widehat{\boldsymbol{q}}$ to the nearest integer and directly passes the gradient during backpropagation. We get
\begin{equation}
    \boldsymbol{a} = D(STE(\widehat{\boldsymbol{q}}))
\end{equation}
The latents are thus trained end-to-end similar to the standard procedure of 3D-GS. Post training, we round $\widehat{\boldsymbol{q}}$ to the nearest integer and use entropy coding for efficiently storing the latents along with the decoder $D$. While each vector in the attribute set $\mathbb{A} = \{\boldsymbol{p},\boldsymbol{s},\boldsymbol{r}, \boldsymbol{c}, o\}$ can be quantized, we do not encode the base band color SH coefficient, the scaling coefficients and the position vector as they are sensitive to initialization and result in large performance drops when quantized. While it is possible to improve feature compression with additional tools such as complex decoders, learnable probability models~\cite{balle2018variational} or Gumbel annealing~\cite{yang2020improving} and so on, they introduce a large overhead in various metrics such as runtime GPU memory and training speed. We aim to utilize an approach which quantizes per-point attributes with little to no cost to these efficiency metrics while still maintaining the reconstruction quality.

\begin{figure}[t]
    \centering
    \begin{tabular}{cc}
         \multirow{2}{*}[7em]{
            \begin{subfigure}{0.6\textwidth}
              \includegraphics[width=\textwidth]{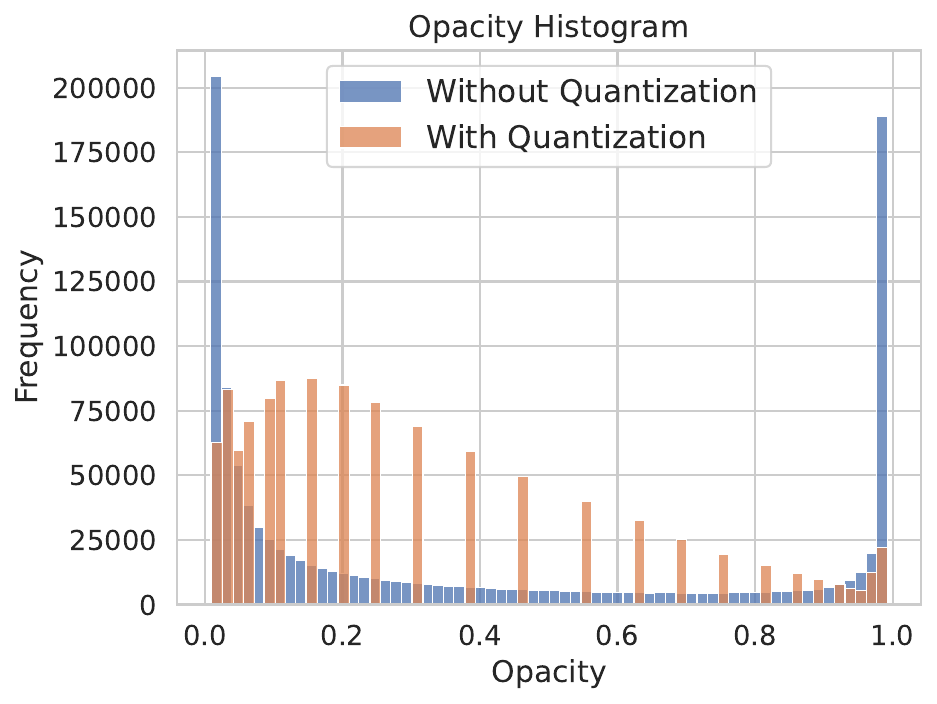}
              \vspace{-1.5em}
              \label{subfig:op_hist}
            \end{subfigure}
        }
         &
            \begin{subfigure}[t]{0.35\textwidth}
              \includegraphics[width=\textwidth]{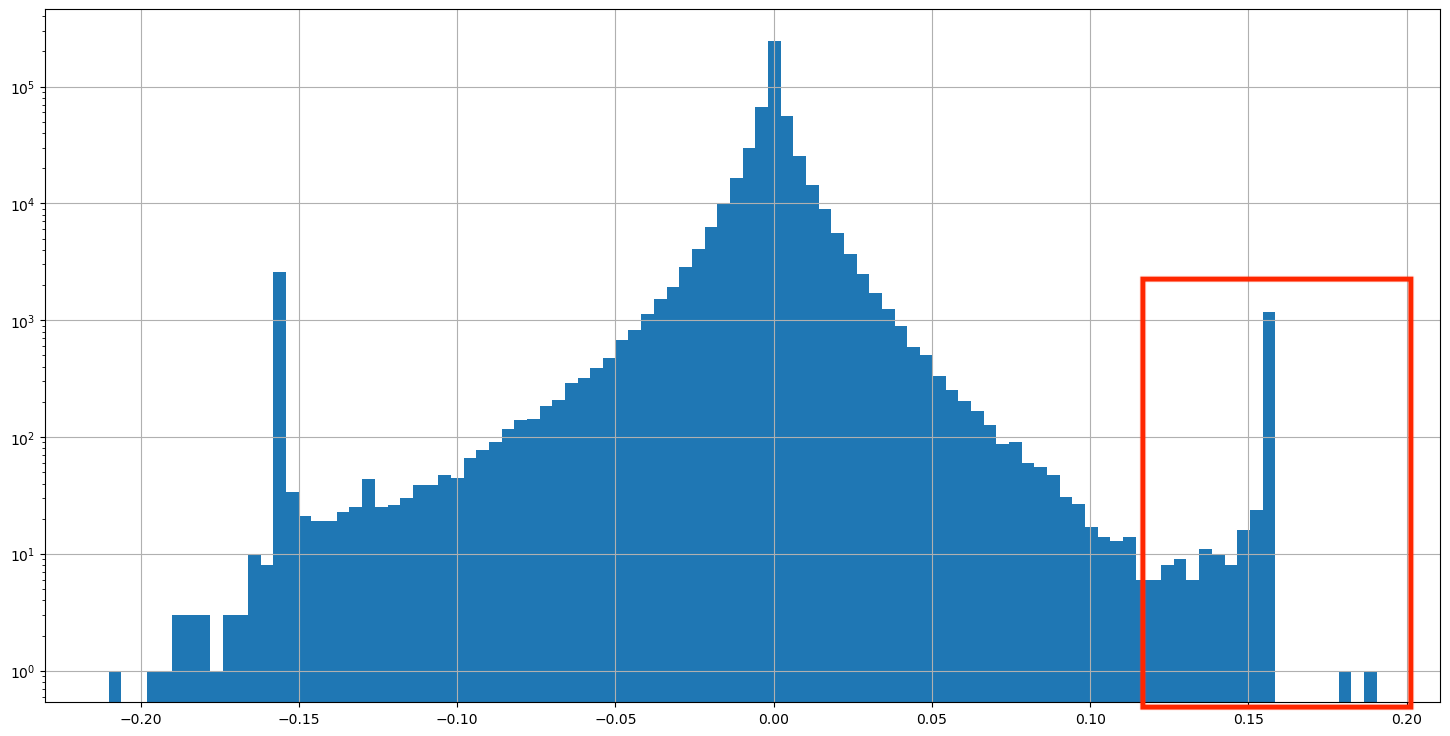}
              \label{subfig:opgrad_gs}
            \end{subfigure}
            \\
         & \begin{subfigure}{0.35\textwidth}
              \includegraphics[width=\textwidth]{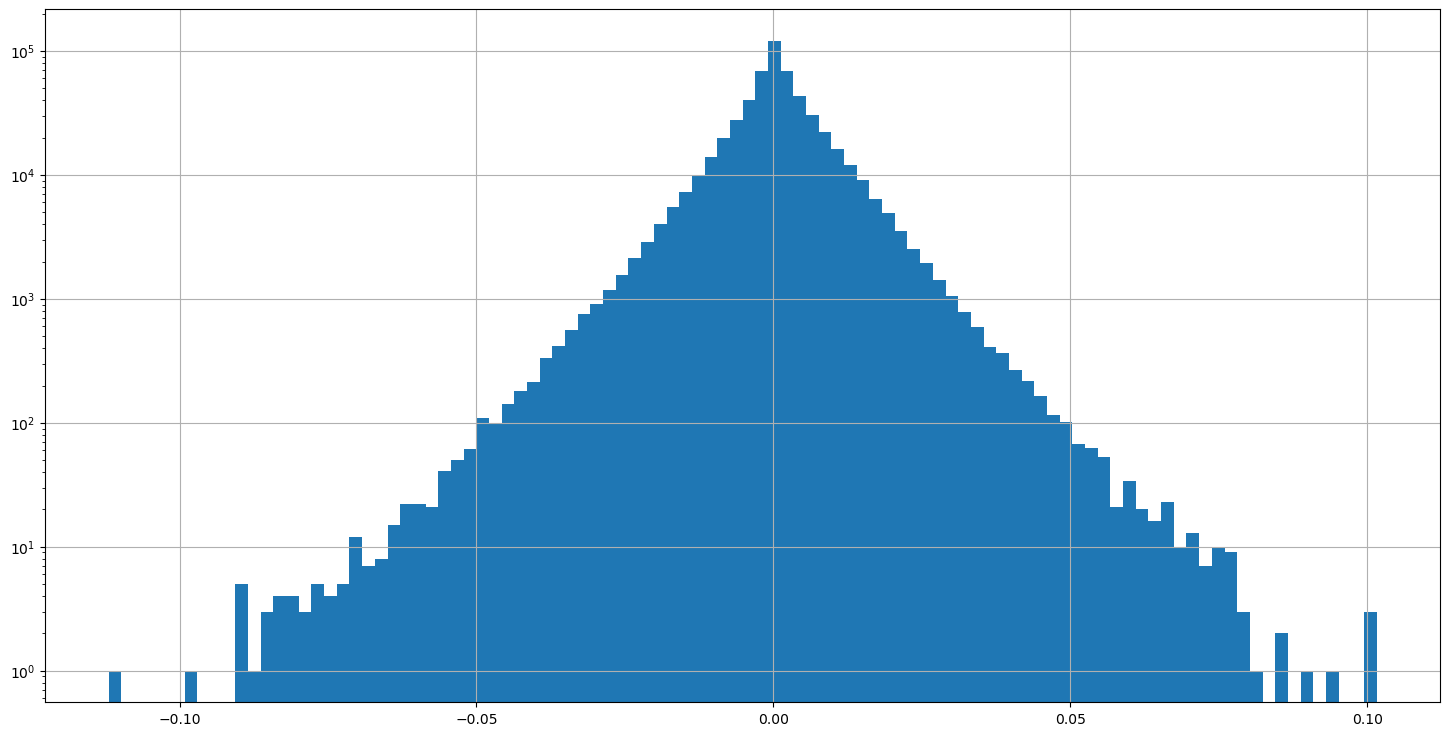}
              \label{subfig:opgrad_ours}
            \end{subfigure}
    \end{tabular}
    \vspace{-1em}
    \caption{Left: Histogram of opacity coefficients with and without quantization. Most coefficients result in values of 0 or 1 without quantization while quantization spreads the opacity values and allows for better blending. Right: Opacity gradient visualization for 3D-GS (top) and \ours (bottom). We reduce outlier gradients while high positive gradients in 3D-GS saturates erroneous Gaussians which are not pruned.}
    \label{fig:opacity_histogram}
    \vspace{-2em}
\end{figure}

\paragraph{Opacity quantization for improved optimization.} While the color and rotation attributes are quantized to reduce the memory footprint, quantizing the opacity coefficients not only reduces memory but improves the optimization process resulting in lesser artifacts in the rendered views. In \cref{fig:opacity_histogram} (left), we visualize the histogram of opacity coefficients of all Gaussian points, with and without quantization. We see that most points converge to 0 or 1 without quantization which is primarily due to large magnitude gradients (top right). While a large negative gradient reduces opacity which can be pruned, a large positive gradient saturates the opacity to 1 leading to artifacts in the rasterization process which is not removed. In contrast, the gradient distribution with quantized opacity coefficients (bottom right) shows fewer outlier gradients and produces a relatively more uniform set of opacities (left). Quantization acts as a soft regularizer as it requires more gradient updates to increase the opacity value from one quantization bin to the next higher bin, thus preventing opacity saturation. In \cref{ssec:ablations}, we show how opacity quantization has the added benefit of removing artifacts normally present in 3D-GS.

\subsection{Progressive training}
\label{ssec:progressive}
Standard training of the Gaussians proceeds by computing the loss over the full image resolution. This results in a more complex loss landscape as the Gaussians are forced to fit to fine features of the scene early in the training. As the SfM initialization is only sparse and several attributes are initialized with rough estimates, the optimization can be suboptimal and result in floating artifacts from Gaussians which cannot be removed during the optimization. We thus propose a coarse-to-fine training strategy by initially rendering at a small scene resolution and gradually increasing the size of the rendered image views over a period of the training iterations until reaching the full resolution. By starting with small images, the Gaussian points easily converge to a good loss minima. This produces better initializations for the creation of further Gaussians through the densification process of cloning and splitting. As the render resolution increases, more Gaussians can be fit to better reconstruct the finer features of the scene. Such a progressive training procedure also helps remove artifacts typically obtained from the rasterization of ill-optimized Gaussians as we show in \cref{ssec:ablations}. This serves as a soft regularization scheme for the creation and deletion of Gaussians. Another added benefit of progressive training is that fewer Gaussians are required to represent coarser scenes while also rendering fewer pixel locations, thereby leading to faster rendering and backpropagation during training. This directly leads to lower training times while still improving the reconstruction quality of the scene upon convergence.

\begin{figure}[t]
\centering
\includegraphics[width=1.0\linewidth]{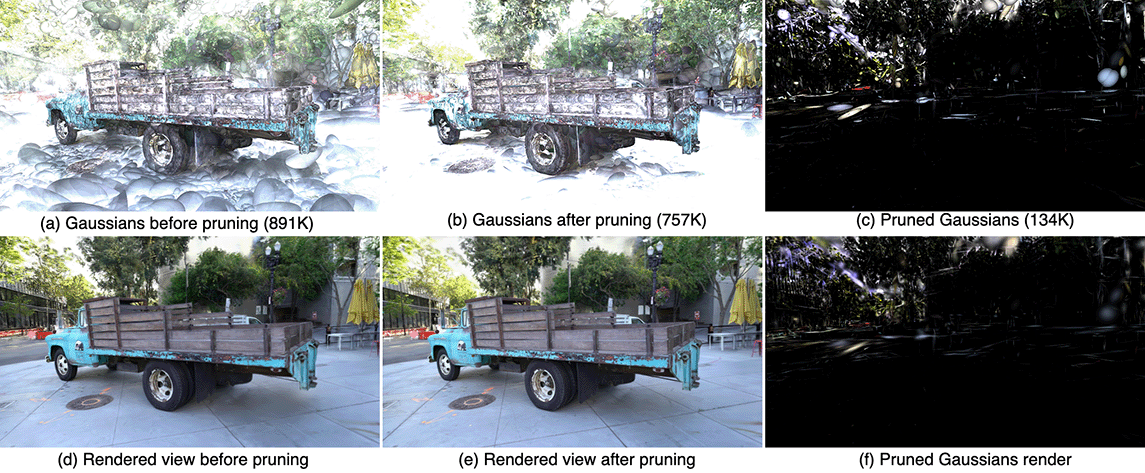}
\vspace{-2em}
\caption{Effect of influence pruning: Our approach identifies Gaussians which do not contribute significantly to the rasterization process and prune them without drop in reconstruction quality.}
\vspace{-1.5em}
\label{fig:abl_pruning}
\end{figure}

\subsection{Influence Pruning}
\label{ssec:densify}
The densification process of cloning and splitting occurs every 100 iterations. This, however, leads to an explosion in the number of Gaussians as a large number of Gaussians exceed the gradient threshold and are either cloned or split. While this can allow for representing finer details in the scene, a significant fraction of the Gaussians are redundant and lead to large training, rendering times as well as memory usage. 3D-GS utilizes an opacity reset stage to remove transparent Gaussians. However, the Gaussian points can have high opacity values while still not influencing the rasterization process due to occlusions (when transmittance T reaches 0 after rendering previous Gaussians in the depth order). The points can also have small scale values influencing very few pixels. To identify the Gaussians most important for reconstructing the full scene, we utilize an influence metric during the rasterization process. More specifically, for the $i^{\textnormal{th}}$ Gaussian to be rendered at pixel location $p$, we define its influence on the pixel and its influence on the full scene as 
\begin{equation}
    W_{\textnormal{i},\textnormal{p}} = \alpha_iT_i = \alpha_i\prod_{j=1}^{i-1}(1-\alpha_j), \quad W_{\textnormal{i}} = \sum_{p} W_{\textnormal{i},\textnormal{p}}
\end{equation}
where $T_i$ measures the transmittance upto Gaussian $i$. We thus obtain a weight vector $\mathbf{W}$, with each element representing the importance of the corresponding Gaussian for rendering in the full scene. Gaussians with small scale values or low opacity values influence fewer pixels and have lower weight values when summed across all pixel locations. Additionally, Gaussians which do not influence the rasterization process ($T_i=0$) have a weight value of zero. This is further visualized in~\cref{fig:abl_pruning}, where for a given view render and a set of Gaussians (left), we obtain nearly identical reconstruction quality (center) with fewer Gaussians. On the right, we visualize the pruned Gaussians which correspond to either highly saturated regions with low transmittance or very small Gaussians with low scale.
This metric has no computational overhead as the weight values are calculated directly during the rasterization process in~\cref{eq:raster}. We thus obtain weight vectors for each iteration and accumulate the weight values over N iterations (set as a hyperparameter) to account for all training views of the full scene. After computing the weight vector, we identify a percentage of the Gaussians with lowest weights and prune them and continue the training process. We show further ablations in \cref{ssec:ablations} to show the effect of the pruning stage in reducing the number of Gaussians while maintaining reconstruction quality. The proposed pruning stage thus removes the Gaussians with the least footprint for scene rendering while the densification process allows for increasing Gaussians to fit to finer scene details.

\section{Experiments}

\subsection{Implementation and evaluation}
We implemented our method by building on \cite{kerbl20233d} which uses a PyTorch framework~\cite{paszke2019pytorch} with a CUDA backend for the rasterization operation. 

A full list of the hyperparameters (learning rates, architecture, initialization of the latents) is provided in the supplementary material. For the progressive scaling, we start with a scale factor of 0.3 while increasing to 1.0 in a cosine schedule. We provide sensitivity analysis of this scale factor in \cref{ssec:ablations}. We perform the scaling schedule for 70\% of the total iterations after which training continues at the full resolution. We fix the opacity reset interval to be every 2500 iterations and the densification frequency to be 175 iterations and the pruning stage for every 5000 iterations until 25000 iterations. At each stage we remove $15\%$ of the Gaussians although a higher value can lead to even larger reductions at the cost of reconstruction quality. We optimize for 30000 iterations but can be controlled based on the time and memory budget for training. We fix the SH degree to be 3 for the color attribute as higher values result in little performance gain for a large increase in memory cost, even with quantization. We use this configuration of hyperparameters for all of our experiments unless mentioned otherwise.

We provide results on 9 scenes from the Mip-Nerf360 dataset~\cite{barron2022mip}, and 2 scenes each from Tanks\&Temples~\cite{knapitsch2017tanks}, Deep Blending~\cite{hedman2018deep} for a total of 13 scenes. These datasets correspond to real-world high resolution scenes which can be unbounded and provide a challenging scenario with parts of the scene scarcely seen during training. We follow the methodology of \cite{kerbl20233d,barron2022mip} with every 8th view used for evaluation and the rest for training. We evaluate the quality of reconstructions primarily with PSNR, and also with the SSIM and LPIPS metrics. We calculate memory of all quantized and non-quantized parameters of the Gaussians for the storage size. The rendering and training memory measures the peak GPU RAM for the full training/rendering phase. We measure the frame rate or Frames Per Second (FPS) based on the time taken to render from all cameras in the scene dataset. Before measuring FPS, we decode all latent attributes using our decoder which is a one-time amortized cost of loading the parameters. For a fair benchmark, the quantitative results comparison of other works in \cref{tab:baselines} are provided by using the numbers reported in \cite{kerbl20233d}, unless mentioned otherwise. The qualitative results are from our own runs of the respective methods.

\subsection{Benchmark comparison}
For NeRFs, we compare against the SOTA method MipNerf360~\cite{barron2022mip} and two recent fast NeRF approaches of INGP~\cite{muller2022instant}, and Plenoxels~\cite{fridovich2022plenoxels}. For our primary baseline, 3D-GS, we provide numbers as reported in \cite{kerbl20233d} and also from our own runs. We show results of our approach for 3 variants: a) training for 30K iterations which is until convergence b) a smaller configuration corresponding to more pruning c) training for 21K iterations which is the end of the progressive training schedule. We summarize the results on all 3 datasets in \Cref{tab:baselines,tab:baselines_db}.
\renewcommand{\arraystretch}{1.6}
\aboverulesep=0pt
\belowrulesep=0pt
\begin{table*}[t]
\caption{Comparison of our approach with prior work in view synthesis on three datasets. * corresponds to our runs on the existing codebase for fair evaluation. We perform competitively in terms of reconstruction metrics while outperforming in terms of efficiency metrics.}
\centering
\resizebox{\linewidth}{!}{
\begin{tabular}{@{}L{\dimexpr.18\linewidth}|
C{\dimexpr.10\linewidth}C{\dimexpr.08\linewidth}C{\dimexpr.08\linewidth}
C{\dimexpr.12\linewidth}C{\dimexpr.06\linewidth}C{\dimexpr.11\linewidth}|
C{\dimexpr.08\linewidth}C{\dimexpr.08\linewidth}C{\dimexpr.08\linewidth}
C{\dimexpr.12\linewidth}C{\dimexpr.06\linewidth}C{\dimexpr.10\linewidth}
@{}}
\toprule
Dataset ($\rightarrow$) & \multicolumn{6}{c|}{Mip-NeRF360}&\multicolumn{6}{c}{Tanks\&Temples}\\
Method & 
\makecell{PSNR\\$\uparrow$} & \makecell{SSIM\\$\uparrow$} & \makecell{LPIPS \\$\downarrow$} & 
\makecell{Storage \\Mem $\downarrow$} & \makecell{FPS \\$\uparrow$} & \makecell{Train \\Time $\downarrow$} & 
\makecell{PSNR\\$\uparrow$} & \makecell{SSIM\\$\uparrow$} & \makecell{LPIPS \\$\downarrow$} & 
\makecell{Storage \\Mem $\downarrow$} & \makecell{FPS \\$\uparrow$} & \makecell{Train \\Time $\downarrow$} \\
\midrule
Plenoxels&
23.08&0.63&0.46&2.1GB&7&25m49s
&21.08&0.72&0.38&2.3GB&13&25m5s\\
INGP&
25.59&0.70&0.33&48MB&9&7m30s
&21.92&0.75&0.31&48MB&14&6m59s\\
M-NeRF360&
27.69&0.79&0.24&9MB&0.06&48h
&22.22&0.76&0.26&9MB&0.14&48h\\
3D-GS&
27.21&0.82&0.21&734MB&134&41m33s&
23.61&0.84&0.18&411MB&154&26m54s\\
3D-GS*&
27.45&0.81&0.22&745MB&110&23m20s& %
23.63&0.85&0.18&430MB&157&12m5s\\ %
\midrule
\ours (Ours)&
27.23&0.81&0.24&54MB&131&21m34s& %
23.37&0.84&0.20&29MB&227&11m39s
\\ %
\cdashline{1-13}
\ours-Small&
26.94&0.80&0.25&47MB&166&17m3s&
23.10&0.82&0.22&19MB&272&10m7s\\ %
\ours-Fast&
26.99&0.81&0.23&71MB&111&16m24s&
23.02&0.83&0.20&38MB&190&8m43s\\
\bottomrule
\end{tabular}
} %
\vspace{-1em}
\label{tab:baselines}
\end{table*}

\begin{figure}[!t]
\noindent\begin{minipage}[t]{0.73\linewidth}%
      \vspace{-1em}
        \captionof{table}{Comparison of our approach on the Deep Blending dataset. We improve reconstruction quality in terms of PSNR compared to 3D-GS while also improving on all efficiency metrics of storage memory, FPS and training time.}
        \label{tab:baselines_db}
        \vspace{0.5em}
        \renewcommand{\arraystretch}{1.4}
        \resizebox{\linewidth}{!}{
        \setlength{\tabcolsep}{5pt}
        \begin{tabular}{@{}L{\dimexpr.27\linewidth}|
        C{\dimexpr.11\linewidth}C{\dimexpr.11\linewidth}C{\dimexpr.11\linewidth}
        C{\dimexpr.13\linewidth}C{\dimexpr.07\linewidth}C{\dimexpr.15\linewidth}
        @{}}
        \toprule
        Dataset ($\rightarrow$) &\multicolumn{6}{c}{Deep Blending}\\
        Method & 
        \makecell{PSNR\\$\uparrow$} & \makecell{SSIM\\$\uparrow$} & \makecell{LPIPS \\$\downarrow$} & 
        \makecell{Storage \\Mem $\downarrow$} & \makecell{FPS \\$\uparrow$}& \makecell{Train \\Time $\downarrow$}  \\
        \midrule
        Plenoxels&
        23.06&0.80&0.51&2.7GB&11&27m49s\\
        INGP&
        24.96&0.82&0.39&48MB&3&8m\\
        M-NeRF360&
        29.40&0.90&0.25&8.6MB&0.09&48h\\
        3D-GS&
        29.41&0.90&0.24&676MB&137&36m2s\\
        3D-GS*&
        29.55&0.90&0.25&656MB&123&23m5s\\ %
        \midrule
        \ours (Ours)&
        29.86&0.91&0.25&52MB&130&21m50s\\ %
        \ours-Small&  %
        29.92&0.90&0.25&33MB&160&17m40s\\
        \ours-Fast&
        29.85&0.91&0.25&63MB&108&16m30s\\
        \bottomrule
        \end{tabular}
        }
    \end{minipage}
    \hspace{0pt}
    \begin{minipage}[t]{0.25\linewidth}%
    \captionof{figure}{3D-GS produces artifacts (right) at various scene locations (arrow) while our rendering produces more scene consistent depth (left). }
    \label{fig:floaters_col}
    \vspace{-1em}
    \includegraphics[width=\linewidth]{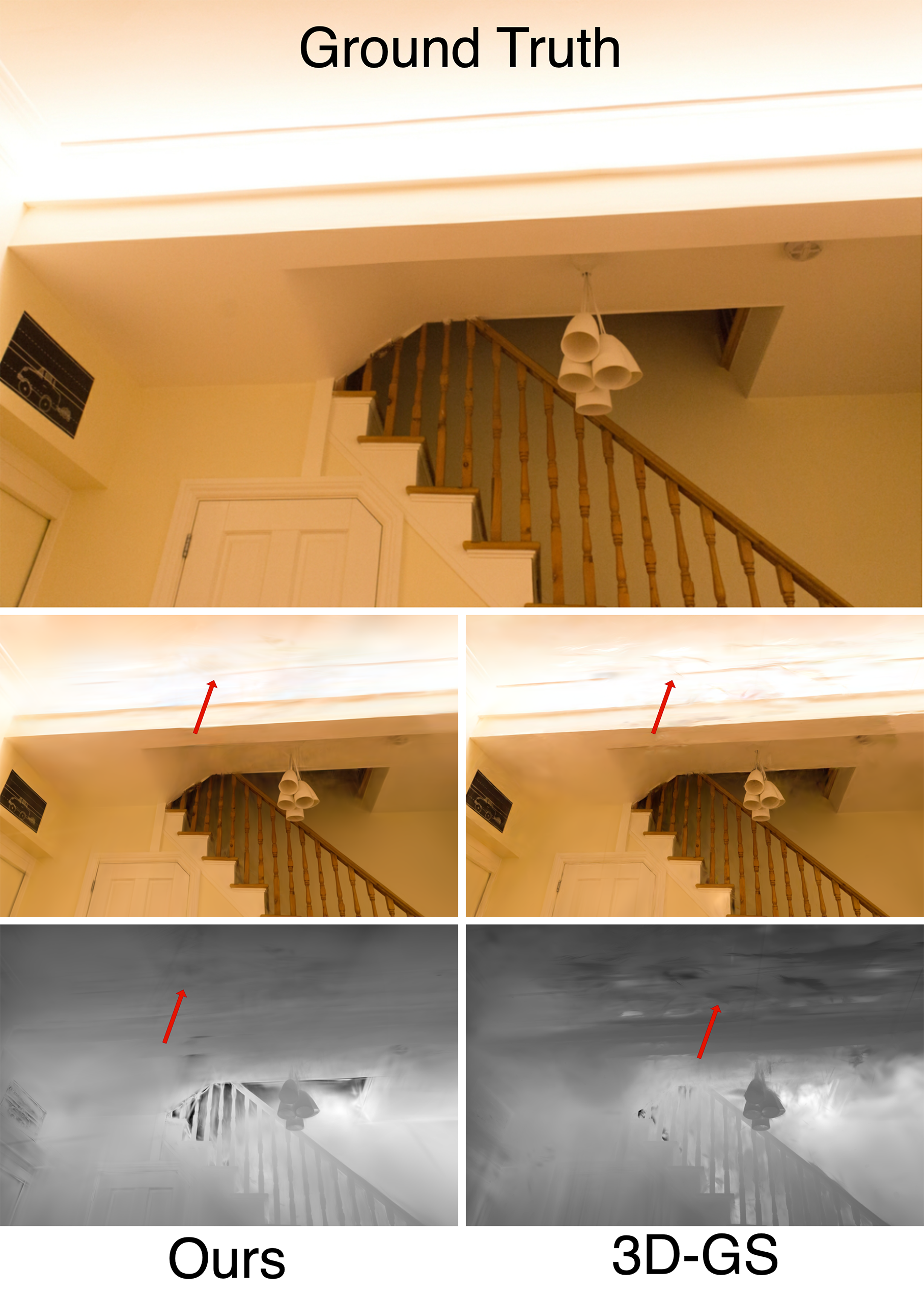}
    \end{minipage}%
    \vspace{-3em}
    \end{figure}
    
\begin{figure*}[t]
    \centering
    \includegraphics[width=\linewidth]{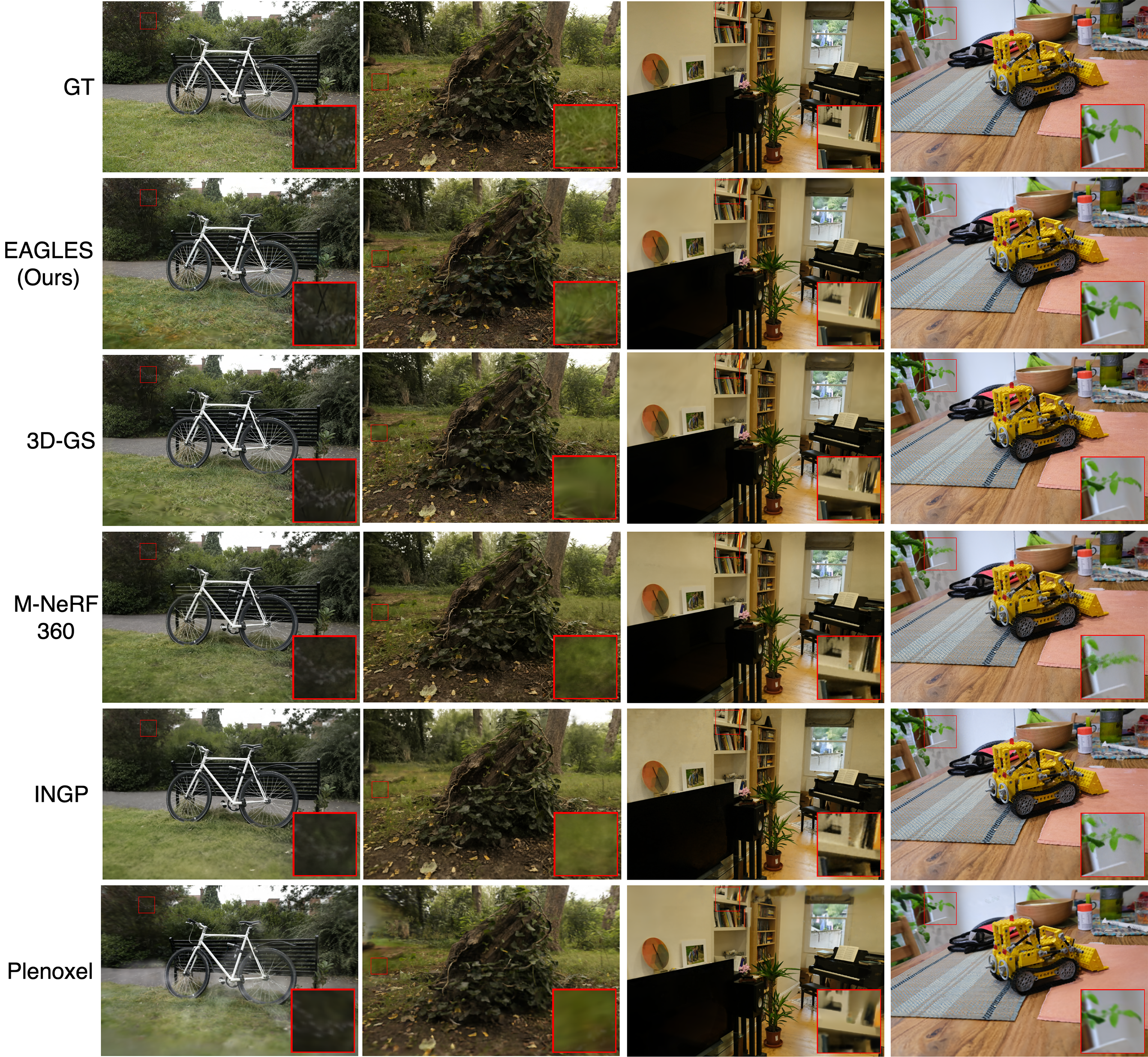}
    \caption{Qualitative comparison of our method with prior work on various scenes. We obtain on par or better reconstructions with the SOTA 3D-GS and MiP-NeRF360 while efficient NeRF methods such as INGP, Plenoxels struggle to reconstruct fine detail. We avoid blurry artifacts at scene boundaries exhibited by 3D-GS and MiP-NeRF360.}
    \label{fig:qualitative}
\end{figure*}

We outperform the voxel-grid based method of Plenoxels on all datasets and metrics. Compared to INGP~\cite{muller2022instant}, another fast Nerf-based method, our approach at 21K iterations, obtains better quality reconstructions at comparable training times on Deep Blending, Tanks\&Temples but higher training times for Mip-NeRF360 dataset. We bridge the gap between NeRF-based methods and Gaussian Splatting in terms of storage memory, by obtaining smaller sizes than INGP (Ours-Small configuration) while still obtaining better reconstruction metrics. We also obtain much higher rendering speeds (${>}15{\times}$) compared to INGP on all datasets paving the way for compact 3D representations with high quality reconstructions and real-time rendering. Against the Mip-NeRF360 approach, we perform competitively in terms of PSNR with a 0.45 dB drop on their dataset and 1.15dB,0.56dB gain on Tanks\&Temples, Deep Blending respectively. While their model is compact in terms of number of parameters they are extremely slow to train (${\sim}48$h) and render (${<}1$FPS). Finally, our reconstructions are on par with 3D-GS achieving minimal performance drops of 0.22dB, 0.26dB PSNR on the Mip-NeRF360, Tanks\&Temples datasets respectively while gaining 0.31dB on Deep Blending. We reduce storage size by  ${\sim}15{\times}$ making the representation suitable for devices with limited memory budgets. Additionally, we accelerate training and rendering compared to 3D-GS obtaining higher FPS and lower train times on all scenes. We additionally see that our approach reaches close to convergence with good visual quality at 21K iterations (marking the end of the progressive scaling period). Note that a fair amount of time is spent on training after 21K iterations due to the full scale render resolution.

We show qualitative results of our approach and other baselines on unseen test views from indoor and outdoor scenes in \cref{fig:qualitative}. Mip-NeRF360 exhibits blurry artifacts such as the grass in the Stump scene (2nd from left) or even incorrect artifacts as seen in the edges of the leaf in Kitchen (right). We obtain reconstructions with quality on-par with 3D-GS or even better reconstructions close to scene boundaries such as the branches in Bicycle (left), grass in Stump (2nd from left). Notably, 3D-GS tends to exhibit numerous floaters at the edges, especially in areas not frequently observed during training. We provide additional visualizations of this in \cref{fig:floaters_col}, showcasing notably smoother reconstructions at scene boundaries, such as the Room's ceiling (3rd from left). This points to a more refined optimization of the point cloud using our approach. We ablate the different components of our approach and analyze the effects of each component in \cref{ssec:ablations} below.

\setlength{\tabcolsep}{4pt}
\begin{table*}[t]
\caption{Ablation of various components of our approach. Attribute quantization significantly reduces the storage memory for a marginal PSNR cost. Progressive training mostly improves the PSNR due to better optimization while the controlled densification and pruning stage signficantly speeds up training and rendering while also reducing storage size at the same or higher PSNR.}
\centering
\resizebox{1.0\linewidth}{!}{
\begin{tabular}{@{}L{\dimexpr.16\linewidth} C{\dimexpr.05\linewidth}C{\dimexpr.08\linewidth}C{\dimexpr.11\linewidth}C{\dimexpr.04\linewidth}
C{\dimexpr.05\linewidth}C{\dimexpr.08\linewidth}C{\dimexpr.11\linewidth}C{\dimexpr.04\linewidth}
C{\dimexpr.05\linewidth}C{\dimexpr.08\linewidth}C{\dimexpr.11\linewidth}C{\dimexpr.04\linewidth}
@{}}
\toprule
\multirow{2}{*}{\quad \ Method}&\multicolumn{4}{c}{Train (Tanks \& Temples)}&\multicolumn{4}{c}{Playroom (Deep Blending)}&\multicolumn{4}{c@{}}{Bicycle (Mip-NeRF360)}\\
\cmidrule(lr){2-5}
\cmidrule(lr){6-9}
\cmidrule(l){10-13}
& PSNR & \makecell{Storage \\Mem } & \makecell{Num. \\Gaussians } & FPS
& PSNR & \makecell{Storage \\Mem } & \makecell{Num. \\Gaussians } & FPS
& PSNR & \makecell{Storage \\Mem } & \makecell{Num. \\Gaussians } & FPS\\
\midrule
Vanilla&\textbf{21.94}&262MB&1.11M&177  %
&30.07&542MB&2.29M&144 %
&\textbf{25.13}&1254MB&5.31M&61\\ %
\makecell[l]{+ Quantization}&21.60&46MB&1.03M&179 %
&\textbf{30.48}&82MB&1.81M&142 %
&24.86&192MB&4.19M&65\\  %
\makecell[l]{+ Progressive}&21.63&38MB&0.85M&194 %
&30.39&75MB&1.67M&140 %
&25.07&190MB&4.19M&71\\ %
+ Densification&21.62&29MB&0.64M&202 %
&30.40&54MB&1.20M&146 %
&25.02&142MB&3.11M&82\\ %
+ Pruning&21.65&\textbf{21MB}&\textbf{0.46M}&\textbf{234} %
&30.38&\textbf{36MB}&\textbf{0.80M}&\textbf{169} %
&25.04&\textbf{104MB}&\textbf{2.26M}&\textbf{87}\\ %
\bottomrule
\end{tabular}
} %
\vspace{-1em}
\label{tab:ablations}
\end{table*}

\subsection{Ablations}
\label{ssec:ablations}

\begin{figure}
    \centering
    \includegraphics[width=\linewidth]{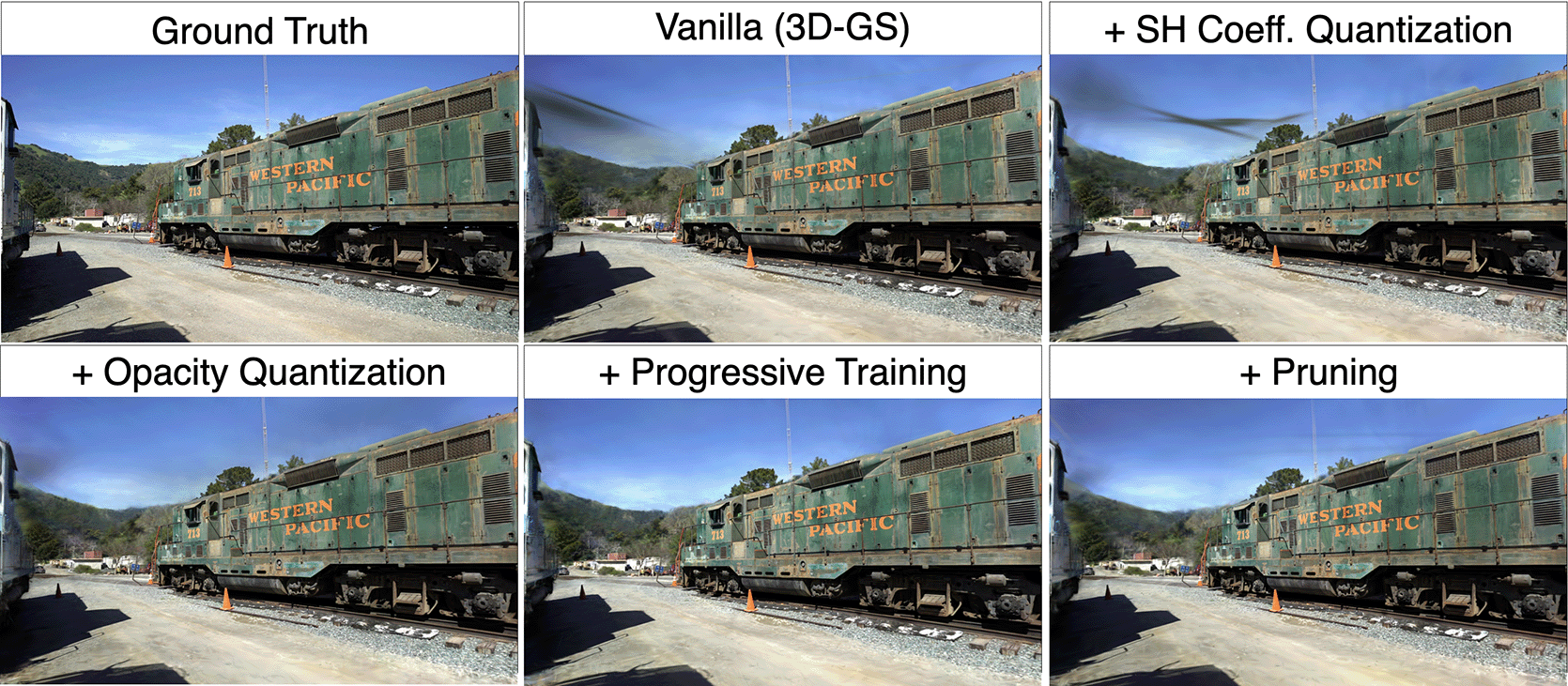}
    \caption{Opacity quantization and progressive retraining removes floaters. Adding the pruning stage does not significantly affect quality while improving efficiency.}
    \label{fig:ablation_qualitative}
    \vspace{-2em}
\end{figure}

For a deeper understanding of our approach, we provide qualitative visualizations for the Train scene and quantitative results for three scenes from each of the 3 datasets, gradually incorporating each component step by step. Results are summarized in \Cref{tab:ablations} and \cref{fig:ablation_qualitative}. "Vanilla" effectively corresponds to the baseline 3D-GS. First, we quantize the color, rotation and opacity attributes for each Gaussian. We get a significant reduction in storage memory with a small drop in PSNR or reconstruction quality. Note that the bulk of the memory post quantization is from the non quantized attributes of scale, position, base color. The quantized attributes are compressed from 220 MB, 452 MB, 1046 MB to 6 MB, 12 MB, 28 MB for the 3 scenes respectively achieving ${\sim}20{-}30{\times}$ memory reduction. We visualize the effect of color and rotation quantization for a single unseen view from the "Train" scene in \cref{fig:ablation_qualitative}. Notice the floaters/rendering artifacts at the top left of the scene as it has little overlap with training views for the vanilla configuration. Quantizing color and rotation does not directly remove these artifacts but opacity quantization significantly improves the visual quality of the rendering as erroneous Gaussians do not saturate quickly.

\begin{figure}[t]
\centering
\setlength{\tabcolsep}{0pt}
\begin{tabular}{cc}
   \includegraphics[width=0.49\linewidth]{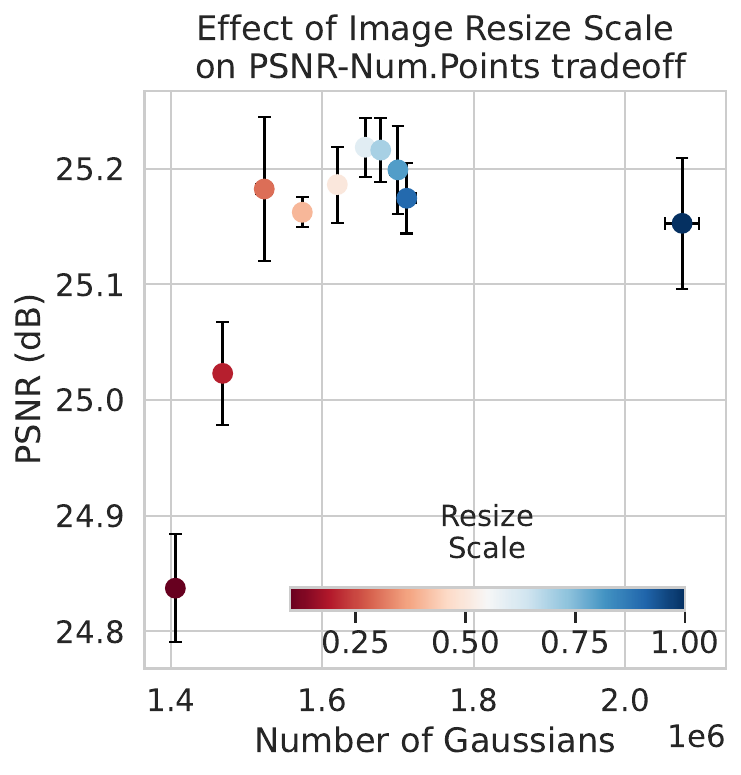}\vspace{-1em}&
    \includegraphics[width=0.49\linewidth]{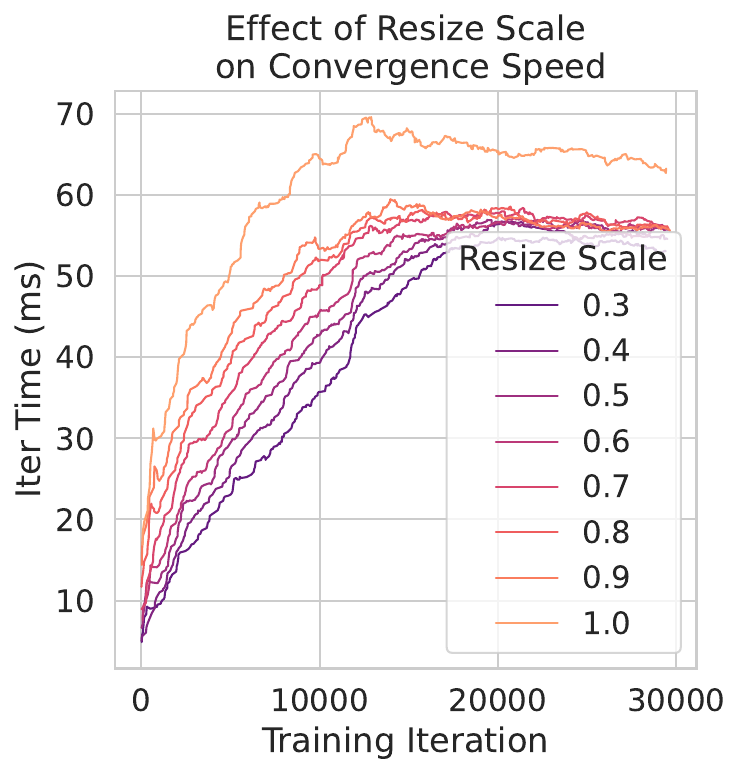}\\
   \quad\quad (a)&\quad\quad(b)
\end{tabular}
\vspace{-1em}
   \caption{Effect of progressive resize scale. (a) Decreasing the scale value leads to reductions in number of Gaussians with no cost to PSNR. (b) This also improves the training speed requiring lesser time per iteration.}
\vspace{-1.5em}
\label{fig:ablations_resize_dense_int}
\end{figure}

We then include progressive scaling increasing the rendering resolution in a cosine schedule. We achieve gains in PSNR with fewer floating artifacts due to a more stable optimization while significantly reducing training time as we show in \cref{tab:ablations_resize}. Progressive scaling also provides a better optimization of the loss landscape removing any remaining foggy artifacts as seen in \cref{fig:ablation_qualitative}. Finally, increasing the densification interval to 175 leads to fewer Gaussians without loss in reconstruction quality (penultimate row). Beyond this value, we observe a sharp drop off in reconstruction quality. However, the
pruning stage continues to decrease the number of Gaussians resulting in lower storage memory, training time, and higher FPS without sacrificing on reconstruction quality in terms of PSNR. This is depicted in the views at \cref{fig:abl_pruning,fig:ablation_qualitative} where the reconstruction quality is similar to without pruning although it reintroduces minor artifacts.

To further analyze the strength of progressive training, we vary the resize scale and visualize the PSNR-model size tradeoff and the convergence speed as well in \cref{fig:ablations_resize_dense_int}(a),(b). We run experiments on the Truck scene and average over 3 random seeds with error intervals reported. From (a), we see that decreasing the scale upto 0.3 has no effect on PSNR but reduces the number of Gaussians needed to represent that scene. Beyond this value, dropoffs in PSNR is observed for lower storage memory. In (b), we analyze the convergence speed in terms of the iteration time over the course of training for various scaling values. As expected, we consistently obtain lower iteration time for lower scale values even with no loss in PSNR as seen in (a).

\subsection{Progressive scaling variants}
\label{ssec:abl_progressive}
As explained previously, progressive scaling of the scene while training provides stable optimizations. We now analyze the effect of applying different types of filters to the image as part of the coarse-to-fine training procedure. Results are summarized in \Cref{tab:ablations_resize}. We try different types of strategies such as a) the mean filter which corresponds to downsampling and re-upsampling the image with bilinear interpolation b) a gaussian filter c) the standard downsampling procedure as used for our experiments and d) with no type of progressive training. For downsampling and mean filtering, we start with a scale of 0.3 and end at 1.0 which corresponds to resizing the image to 30\% of its dimensions and scaling upto its original size gradually for a period of 70\% of the iterations. For Gaussian filtering, we progressively decrease the filter size from the initial value specified in the table down to $1\times1$, which essentially equates to no filtering. Compared to the no filter case, all other types of filters result in fewer Gaussians leading to lower memory, training time and higher FPS. Both Gaussian and Mean filters provide large gains in terms of efficiency metrics with little to no drops in PSNR. The Gaussian filter naturally provides a coarse-to-fine schedule for training Gaussian points. Nonetheless, the training still proceeds at full resolution and the largest gains in terms of training time is produced with downsampling. The $5\times5$ Gaussian filter produces similar results as downsampling albeit with higher training times while we observe a larger Gaussian filter $15\times 15$ leads to much higher efficiency at the cost of PSNR.

\setlength{\tabcolsep}{0pt}
 \renewcommand{\arraystretch}{1.2}
\begin{table}[t]
\caption{Various types of progressive scaling. Downsampling reduces training time the most while high Gaussian filter sizes improves size and FPS at the cost of reconstruction quality.}
\centering
\resizebox{0.9\linewidth}{!}{
\begin{tabular}{@{}L{\dimexpr.25\linewidth} C{\dimexpr.12\linewidth}C{\dimexpr.14\linewidth}C{\dimexpr.14\linewidth}C{\dimexpr.11\linewidth}C{\dimexpr.12\linewidth}@{}}
\toprule
\makecell[l]{Filter Type}& PSNR & \makecell{Storage \\Mem } & \makecell{Num. \\Gaussians } & FPS&\makecell{Training\\Time}\\
\midrule
None&23.34dB&43MB&0.95M&211&13m27s\\
Mean&23.31dB&27MB&0.61M&280&11m41s\\
Gaussian $(5{\times}5)$&23.41dB&34MB&0.74M&248&12m8s\\
Gaussian $(7{\times}7)$&23.36dB&28MB&0.61M&276&11m32s\\
Gaussian $(15{\times}15)$&23.17dB&\textbf{21MB}&\textbf{0.46M}&\textbf{321}&10m51s\\ 
\makecell[l]{Downsample}&\textbf{23.41dB}&34MB&0.75M&244&\textbf{9m49s}\\
\bottomrule
\end{tabular}
} %
\label{tab:ablations_resize}
\end{table}

 \renewcommand{\arraystretch}{1.2}
\begin{table}[t]
\caption{We require much lesser training and rendering memory consumption by ours and 3D-GS across all scenes consistently.}
\vspace{-0.5em}
\centering
\resizebox{0.8\linewidth}{!}{
\begin{tabular}{@{}L{\dimexpr.1\linewidth} C{\dimexpr.11\linewidth}C{\dimexpr.11\linewidth}
C{\dimexpr.11\linewidth}C{\dimexpr.11\linewidth}
C{\dimexpr.11\linewidth}C{\dimexpr.11\linewidth}
@{}}
\toprule
\multirow{2}{*}{\makecell[l]{Method}}& \multicolumn{2}{c}{Bicycle} & \multicolumn{2}{c}{Truck} & \multicolumn{2}{c}{Playroom} \\
\cmidrule(lr){2-3} \cmidrule(lr){4-5} \cmidrule(l){6-7} 
& Train & Render & Train & Render & Train &Render\\
\midrule
3D-GS & 17.4G&9.5G&8.5G&4.8G&9.6G&6.0G\\
\ours & 10G& 7.4G&5.3G&3.6G&7.1G&5.3G\\
\bottomrule
\end{tabular}
} %
\vspace{-1em}
\label{tab:compute_mem}
\end{table}

\subsection{Training and Rendering Memory}
In this section, we show the memory consumption of our approach and 3D-GS on the 3 datasets in \Cref{tab:compute_mem}. We measure peak GPU memory used during the training or rendering phase by our approach and 3D-GS. We see that we require much lesser memory during training even with latents and decoders. Since our quantization decodes the latents to floating point values before a forward or backward pass, no gains are obtained in terms of runtime memory consumption for each Gaussian. However, with progressive training and the pruning stage, we obtain significantly lower number of Gaussians leading to lower runtime memory during training/rendering. For the Bicycle scene especially, compared to the 17G required by 3D-G, we consume only 10G GPU RAM during training making it practical for many consumer GPUs with 12G RAM.

\section{Conclusion}

In this work, we proposed a simple yet powerful approach for 3D reconstruction and novel view synthesis. We build upon the seminal work on 3D Gaussian Splatting~\cite{kerbl20233d}, and propose major improvements that not only reduces the storage requirements for each scene by 10-20$\times$, but also achieves it with lower training cost, faster inference time, and on par reconstruction quality. We achieve this by 3 major improvements over the prior work - attribute quantization for per-point compression, progressive training for faster training and better reconstruction, and a pruning stage for reducing number of points for the scene representation. Our extensive quantitative and qualitative analyses shows the efficacy of our approach in 3D representation.

\noindent
\textbf{Acknowledgements:} \small{This work was partially supported by IARPA via Department of Interior/Interior Business Center (DOI/IBC) contract number 140D0423C0076. The U.S. Government is authorized to reproduce and distribute reprints for Governmental purposes notwithstanding any copyright annotation thereon. The authors acknowledge UMD’s supercomputing resources made available for conducting this research. The views and conclusions contained herein are those of the authors and should not be interpreted as necessarily representing the official policies or endorsements, either expressed or implied, of IARPA, DOI/IBC, or the U.S. Government. We also thank Jon Barron for providing additional scenes from the Mip-NeRF360 dataset for our experiments.}

\bibliographystyle{splncs04}
\bibliography{main}

\title{Supplementary - EAGLES: Efficient Accelerated 3D Gaussians with Lightweight EncodingS}

\author{Sharath Girish\inst{1} \and
Kamal Gupta\inst{2} \and
Abhinav Shrivastava\inst{3}
}

\authorrunning{S.~Girish et al.}

\institute{\email{sgirish@cs.umd.edu}\and
\email{kamalgupta308@gmail.com}\and
\email{abhinav@cs.umd.edu}\\
University of Maryland, College Park
}
\maketitle

\section{Hyperparameters}
\label{supp_sec:latent_hyperparam}
We compress the color, rotation and opacity attributes of each Gaussian as explained in the main paper. Each attribute consists of several hyperparameters; mainly latent dimension, decoder parameter learning rate, latent learning rate, decoder initialization. The decoder parameters are initialized using a normal distribution with a standard deviation. As the uncompressed attributes $\boldsymbol{a}$ are initialized using SfM for 3D-GS~\cite{kerbl20233d}, we obtain the latent initialization (with continuous approximations $\widehat{\boldsymbol{q}}$) by inverting the decoder $D$. 
\begin{equation}
    \widehat{\boldsymbol{q}} = D^{-1}(\boldsymbol{a})
\end{equation}
For a decoder which is only a linear layer, a least square approximation provides the latent values. The learning rate of the latents is obtained by scaling the original attribute learning rate with a scale factor and divided by the norm of the decoder (for a linear layer). This improves training stability and convergence when decoder norm is either too high or too low. Values used for all the compressible attributes are provided in \cref{supp_tab:latent_hyper}. We use these values for all of our experiments and find it to be stable across various datasets. All other hyperaparameter values are used as is the default in~\cite{kerbl20233d}. 

\begin{table}[t]
\caption{Latent hyperparameter values.}
\vspace{-0.5em}
\centering
\resizebox{1.0\linewidth}{!}{
\begin{tabular}{@{}
C{\dimexpr.2\linewidth}C{\dimexpr.2\linewidth}
C{\dimexpr.2\linewidth}C{\dimexpr.2\linewidth}
C{\dimexpr.16\linewidth}@{}}
\toprule
Attribute &\makecell{Latent\\Dimension}&\makecell{Decoder\\LR}&\makecell{Decoder\\Std.}&\makecell{Latent\\LR Scale}\\
\midrule
Color&16&0.0001&0.0005&1.0\\
Rotation&8&0.0001&0.01&1.0\\
Opacity&1&0.0001&0.5&1.0\\
\bottomrule
\end{tabular}
} %
\vspace{-1em}
\label{supp_tab:latent_hyper}
\end{table}

\section{Per scene metrics}
We provide metrics for each scene across the 3 datasets of Mip-NeRF360, Tanks\&Temples, and Deep Blending in \Cref{supp_tab:mip}, \Cref{supp_tab:tandt}, \Cref{supp_tab:db} respectively.

\begin{table*}[t]
\caption{MiP-NeRF360 per scene results}
\vspace{-0.5em}
\centering
\resizebox{1.0\linewidth}{!}{
\begin{tabular}{@{}
C{\dimexpr.1\linewidth}C{\dimexpr.1\linewidth}
C{\dimexpr.1\linewidth}C{\dimexpr.1\linewidth}
C{\dimexpr.1\linewidth}C{\dimexpr.1\linewidth}
C{\dimexpr.1\linewidth}C{\dimexpr.1\linewidth}
C{\dimexpr.1\linewidth}
@{}}
\toprule
Scene & Method & PSNR & SSIM & LPIPS & \makecell{Storage\\Mem}&FPS &\makecell{Train\\Time}&\makecell{Num.\\Gaussians}\\
\midrule
\multirow{2}{*}{Bicycle} & Ours & 25.04 & \textbf{0.75} & 0.24 & \textbf{104MB} & \textbf{87} & \textbf{24m 53s} & \textbf{2.26M} \\
& 3D-GS & \textbf{25.13} & 0.75 & \textbf{0.24} & 1254MB & 61 & 28m 44s & 5.31M \\
\midrule
\multirow{2}{*}{Bonsai} & Ours & 31.32 & 0.94 & 0.19 & \textbf{29MB} & 177 & \textbf{17m 9s} & \textbf{0.64M} \\
& 3D-GS & \textbf{32.19} & \textbf{0.95} & \textbf{0.18} & 295MB & \textbf{187} & 18m 11s & 1.25M \\
\midrule
\multirow{2}{*}{Counter} & Ours & 28.40 & 0.90 & 0.20 & \textbf{25MB} & 138 & \textbf{19m 55s} & \textbf{0.56M} \\
& 3D-GS & \textbf{29.11} & \textbf{0.91} & \textbf{0.18} & 276MB & \textbf{139} & 21m 29s & 1.17M \\
\midrule
\multirow{2}{*}{Flowers} & Ours & 21.29 & 0.58 & 0.37 & \textbf{60MB} & \textbf{144} & \textbf{18m 48s} & \textbf{1.33M} \\
& 3D-GS & \textbf{21.37} & \textbf{0.59} & \textbf{0.36} & 818MB & 105 & 22m 14s & 3.47M \\
\midrule
\multirow{2}{*}{Garden} & Ours & 26.91 & 0.84 & 0.15 & \textbf{74MB} & \textbf{119} & \textbf{23m 7s} & \textbf{1.65M} \\
& 3D-GS & \textbf{27.32} & \textbf{0.86} & \textbf{0.12} & 1343MB & 65 & 28m 57s & 5.69M \\
\midrule
\multirow{2}{*}{Kitchen} & Ours & 30.77 & 0.93 & 0.13 & \textbf{45MB} & \textbf{116} & 25m 5s & \textbf{1.00M} \\
& 3D-GS & \textbf{31.53} & \textbf{0.93} & \textbf{0.12} & 417MB & 109 & \textbf{24m 57s} & 1.77M \\
\midrule
\multirow{2}{*}{Room} & Ours & 31.47 & 0.92 & 0.20 & \textbf{30MB} & 123 & 21m 38s & \textbf{0.67M} \\
& 3D-GS & \textbf{31.59} & \textbf{0.92} & \textbf{0.20} & 353MB & \textbf{131} & \textbf{21m 37s} & 1.50M \\
\midrule
\multirow{2}{*}{Stump} & Ours & \textbf{26.78} & \textbf{0.77} & \textbf{0.24} & \textbf{100MB} & \textbf{128} & \textbf{20m 2s} & \textbf{2.22M} \\
& 3D-GS & 26.73 & 0.77 & 0.24 & 1042MB & 97 & 22m 8s & 4.42M \\
\midrule
\multirow{2}{*}{Treehill} & Ours & \textbf{22.69} & \textbf{0.64} & \textbf{0.34} & \textbf{72MB} & \textbf{129} & 21m 49s & \textbf{1.60M} \\
& 3D-GS & 22.61 & 0.64 & 0.35 & 807MB & 102 & \textbf{21m 46s} & 3.42M \\
\midrule
\multirow{2}{*}{Average} & Ours & 27.23 & 0.81 & 0.24 & \textbf{54MB} & \textbf{131} & \textbf{21m 34s} & \textbf{1.33M} \\
& 3D-GS & \textbf{27.45} & \textbf{0.81} & \textbf{0.22} & 745MB & 110 & 23m 20s & 3.11M \\
\bottomrule
\end{tabular}
} %
\vspace{-1em}
\label{supp_tab:mip}
\end{table*}

\begin{table*}[t]
\caption{Tanks\&Temples per scene results}
\vspace{-0.5em}
\centering
\resizebox{1.0\linewidth}{!}{
\begin{tabular}{@{}
C{\dimexpr.1\linewidth}C{\dimexpr.1\linewidth}
C{\dimexpr.1\linewidth}C{\dimexpr.1\linewidth}
C{\dimexpr.1\linewidth}C{\dimexpr.1\linewidth}
C{\dimexpr.1\linewidth}C{\dimexpr.1\linewidth}
C{\dimexpr.1\linewidth}
@{}}
\toprule
Scene & Method & PSNR & SSIM & LPIPS & \makecell{Storage\\Mem}&FPS &\makecell{Train\\Time}&\makecell{Num.\\Gaussians}\\
\midrule
\multirow{2}{*}{Train} & Ours & 21.65 & 0.80 & 0.24 & \textbf{21MB} & \textbf{234} & \textbf{11m 27s} & \textbf{0.46M} \\
& 3D-GS & \textbf{21.94} & \textbf{0.81} & \textbf{0.20} & 262MB & 177 & 11m 43s & 1.11M \\
\midrule
\multirow{2}{*}{Truck} & Ours & 25.09 & 0.87 & 0.16 & \textbf{38MB} & \textbf{220} & \textbf{11m 50s} & \textbf{0.83M} \\
& 3D-GS & \textbf{25.31} & \textbf{0.88} & \textbf{0.15} & 599MB & 139 & 12m 27s & 2.54M \\
\midrule
\multirow{2}{*}{Average} & Ours & 23.37 & 0.84 & 0.20 & \textbf{29MB} & \textbf{227} & \textbf{11m 39s} & \textbf{0.65M} \\
& 3D-GS & \textbf{23.63} & \textbf{0.85} & \textbf{0.18} & 430MB & 157 & 12m 5s & 1.83M \\
\bottomrule
\end{tabular}
} %
\vspace{-1em}
\label{supp_tab:tandt}
\end{table*}

\begin{table*}[t]
\caption{Deep Blending per scene results}
\vspace{-0.5em}
\centering
\resizebox{1.0\linewidth}{!}{
\begin{tabular}{@{}
C{\dimexpr.1\linewidth}C{\dimexpr.1\linewidth}
C{\dimexpr.1\linewidth}C{\dimexpr.1\linewidth}
C{\dimexpr.1\linewidth}C{\dimexpr.1\linewidth}
C{\dimexpr.1\linewidth}C{\dimexpr.1\linewidth}
C{\dimexpr.1\linewidth}
@{}}
\toprule
Scene & Method & PSNR & SSIM & LPIPS & \makecell{Storage\\Mem}&FPS &\makecell{Train\\Time}&\makecell{Num.\\Gaussians}\\
\midrule
\multirow{2}{*}{Drjohnson} & Ours & \textbf{29.35} & \textbf{0.90} & \textbf{0.24} & \textbf{69MB} & 92 & \textbf{25m 47s} & \textbf{1.57M} \\
& 3D-GS & 28.77 & 0.90 & 0.25 & 769MB & \textbf{102} & 25m 9s & 3.26M \\
\midrule
\multirow{2}{*}{Playroom} & Ours & \textbf{30.38} & \textbf{0.91} & 0.25 & \textbf{36MB} & \textbf{169} & \textbf{17m 52s} & \textbf{0.80M} \\
& 3D-GS & 30.07 & 0.90 & \textbf{0.25} & 542MB & 144 & 21m 0s & 2.29M \\
\midrule
\multirow{2}{*}{Average} & Ours & \textbf{29.86} & \textbf{0.91} & \textbf{0.25} & \textbf{52MB} & \textbf{130} & \textbf{21m 50s} & \textbf{1.19M} \\
& 3D-GS & 29.42 & 0.90 & 0.25 & 656MB & 123 & 23m 5s & 2.78M \\
\bottomrule
\end{tabular}
} %
\vspace{-1em}
\label{supp_tab:db}
\end{table*}

\end{document}